\documentclass{IEEEtran}
\usepackage{mathtools}
\usepackage{comment}

\usepackage{color}
\usepackage{paralist}
\usepackage{subfigure}
\usepackage{stackengine}
\usepackage{adjustbox}
\usepackage{amsmath,amssymb,amsfonts}
\usepackage{multirow}
\usepackage{url}
\usepackage{diagbox}
\usepackage{mathtools}

\DeclarePairedDelimiter\floor{\lfloor}{\rfloor}

\begin{document}
\title{Counting with Adaptive Auxiliary Learning}

\author{Yanda~Meng, Joshua~Bridge, Meng~Wei, Yitian~Zhao, Yihong~Qiao, Xiaoyun~Yang, Xiaowei~Huang, and Yalin~Zheng* 
\thanks{Y. Meng, J. Bridge and Y. Zheng are with the Department of Eye and Vision Science, University of Liverpool, Liverpool, L7 8TX, United Kingdom.

M. Wei was with the Department of Eye and Vision Science, University of Liverpool, Liverpool, L7 8TX, United Kingdom.

Y. Zhao is with the Cixi Institute of Biomedical Engineering, Ningbo Institute of Materials Technology and Engineering, Chinese Academy of Science, Ningbo 315201, China.

Y. Qiao is with the China Science IntelliCloud Technology Co., Ltd, Shanghai, China.

X. Yang is with Remark AI UK Limited, London, SE1 9PD, United Kingdom.

X. Huang is with the Department of Computer Science, University of Liverpool, Liverpool, L69 3BX, United Kingdom.

Corresponding author: Yalin Zheng (yalin.zheng@liverpool.ac.uk).}
}

\markboth{}%
{Shell \MakeLowercase{\textit{et al.}}: Bare Demo of IEEEtran.cls for IEEE Journals}

\maketitle

\begin{abstract}
This paper proposes an adaptive auxiliary task learning based approach for object counting problems. Unlike existing auxiliary task learning based methods, we develop an attention-enhanced adaptively shared backbone network to enable both task-shared and task-tailored features learning in an end-to-end manner.
The network seamlessly combines standard Convolution Neural Network (\textit{CNN}) and Graph Convolution Network (\textit{GCN}) for feature extraction and feature reasoning among different domains of tasks.
Our approach gains enriched contextual information by iteratively and hierarchically fusing the features across different task branches of the adaptive \textit{CNN} backbone. The whole framework pays special attention to the objects' spatial locations and varied density levels, informed by object (or crowd) segmentation and density level segmentation auxiliary tasks.
In particular, thanks to the proposed dilated contrastive density loss function, our network benefits from individual and regional context supervision in terms of pixel-independent and pixel-dependent feature learning mechanisms, along with strengthened robustness.
Experiments on seven challenging multi-domain datasets demonstrate that our method achieves superior performance to the state-of-the-art auxiliary task learning based counting methods. Our code is made publicly available at: 
\url{https://github.com/smallmax00/Counting_With_Adaptive_Auxiliary}
\end{abstract}

\begin{IEEEkeywords}
Objects Counting, \textit{GCN}, Dilated Contrastive Density Loss, Adaptive Auxiliary Task
\end{IEEEkeywords}

%
\IEEEpeerreviewmaketitle

\section{Introduction}
\IEEEPARstart{O}{bject} counting by inferring the number of objects in images or video contents is a crucial yet challenging computer vision task.
This paper is primarily motivated to address crowd counting problems while it can be applied to other counting problems such as cell and vehicle counting. Due to the need for crowd gathering in many scenarios such as parades, concerts, and stadiums, a robust and accurate crowd counting model plays an essential role in multimedia applications for security alerts, public space design, crowd management, \(etc.\) \cite{gao2020cnn}. 

\begin{figure}
\centering
\includegraphics[width=9cm]{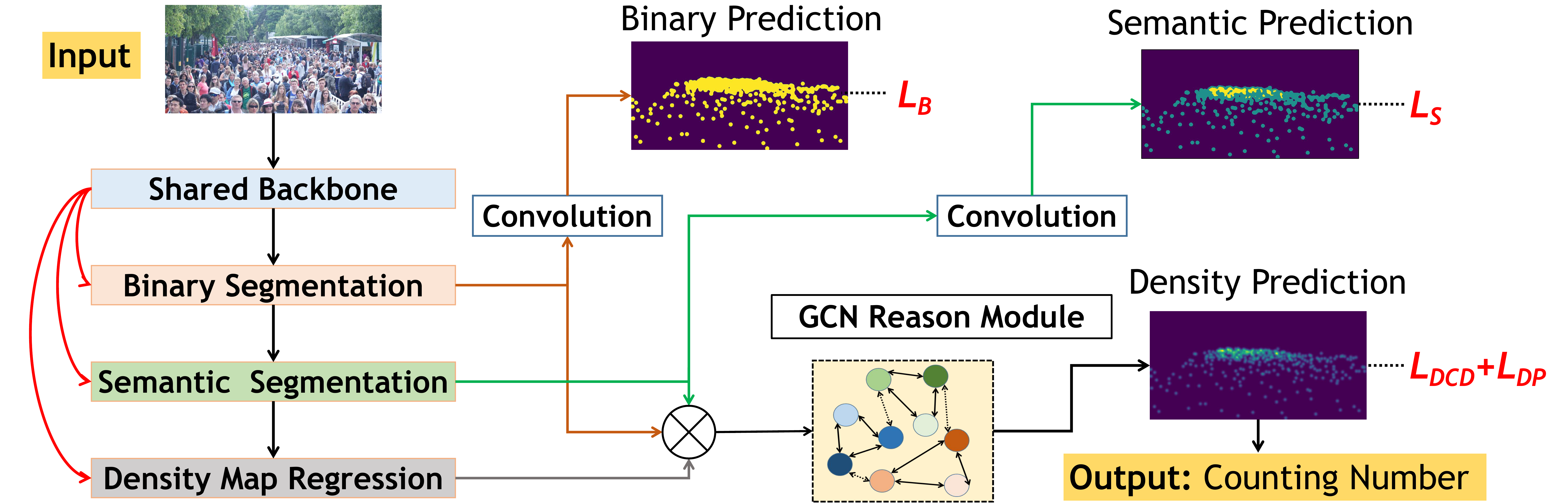} 
\caption{Overview of the proposed network structure in the scene of crowd counting. An attention-enhanced adaptively shared backbone network is proposed to enable both task-shared and task-tailored features learning. A novel Graph Convolution Network (\textit{GCN}) reasoning module is introduced to tackle issues of cross-domain features reasoning among three different tasks. A novel loss function \(L_{DCD}\) is proposed to take into account more adjacent pixels for regional density difference, which strengthens the network's generalizability.}
\label{overview}
\end{figure}

Benefits from Convolutional Neural Network (\textit{CNN})'s excellent feature learning ability, the performance of crowd counting approaches has consistently been improved. Recent state-of-the-art approaches, such as \cite{onoro2016towards,guo2019sau,reddy2020adacrowd,meng2021spatial,liu2020efficient}, showed that a density map regression paradigm gains satisfying results. In these methods, given an input image, a \textit{CNN}-based network is used to regress the corresponding density map; the summed pixel value of the density map gives the total counting numbers in that image. However, there is still much room to improve the counting performance due to challenging issues \cite{gao2020cnn},  such as significant scale changes, wide variations of density levels, and complex scene backgrounds. To solve these problems, some previous methods \cite{shi2019counting,liu2018decidenet,jiang2020density,modolo2021understanding,wan2019residual} relied on different types of information granularity in terms of `auxiliary task learning'. 
These methods applied a single shared backbone network structure to extract generalized features for all the tasks. Unfortunately, this strategy may lead to under-fitting as the generalizable representation often cannot effectively describe the comprehensive cross-domain features across different tasks at the same time \cite{liu2019end,gao2020feature}. Intuitively, our motivation is that the backbone network should be able to yield both generic (or universal) representations shareable for all the tasks and specialized features tailored for individual tasks. To this end, we designed an attention-based adaptive shareable backbone network to enable task-shared and task-customized features learning in an end-to-end manner (Fig. \ref{overview} shows an overview of the network architecture). Note that, the term `auxiliary task learning' is referred to as the feature learning of different density information granularity levels. Specifically, the crowd segmentation task and the density level segmentation task in Fig. \ref{overview} are the auxiliary tasks, and the density map regression task is the main task. We generated the ground truth of crowd segmentation and density level segmentation tasks from the density map regression task's ground truth. Intuitively, the information from the ground truth of auxiliary tasks is not increased; however, the information is enhanced and specified through auxiliary tasks in terms of different density information granularity.

Instead of aiming to achieve the best performance for all the tasks, our adaptive shareable backbone network primarily focused on optimizing the primary density map regression task, along with the multi-granularity information enhancement from the other auxiliary tasks. Our backbone network contained a multi-level information aggregation mechanism to iteratively and hierarchically fuse features learned from different stages and auxiliary branches to tackle large scale changes problems. 
We also applied two attention-based auxiliary task learning branches: 1.) crowd segmentation task to indicate spatial regions of interest to tackle complex background issues; and 2.) density level segmentation task to be aware of the varied density levels across the image, which can tackle the significant variations of density problems. An ablation study demonstrates that our adaptive auxiliary branches improve performance over a single-shared backbone based auxiliary task learning network with comparable model size.

Apart from the aforementioned components, we also studied how to reason and fuse features from different tasks for density map regression. The features extracted from crowd segmentation and density level segmentation branches belong to different feature domains with various granularity representations. Direct fusion (\textit{e.g.} element-wise multiplication or channel-wise concatenation) among three task branches' outputs can result in domain conflicts \cite{luo2020hybrid}. Thus, further reasoning is necessary to improve the counting performance. To this end, we exploited the information reasoning nature of Graph Convolutional Networks (\textit{GCN}). \textit{GCN} has recently shown promising reasoning ability on many computer vision tasks, such as scene understanding \cite{li2018beyond}, image segmentation \cite{Meng2020GCN,meng2020cnn},\textit{ etc.}, but has been rarely studied in the crowd counting task domain. 
Our model projects a collection of spatial-aware density feature map's pixels with similar density levels to each graph vertex and exploits a \textit{GCN} to reason about the relations among graph vertices. This is different from a recent work \cite{luo2020hybrid}, which directly treated cross-domain feature maps as graph vertices and utilized a cascaded Graph Neural Network (GNN) to reason the cross-scale relationships. 
Our experiment results have proved that the proposed \textit{GCN} reasoning module helps to improve the counting accuracy.

We also proposed a novel loss function to supervise the main task learning processes. Notably, for the supervision of density map regression, the widely adopted Least Absolute Error \textit{(L1)} or Least Square Error \textit{(L2)} loss in previous counting methods \cite{zhang2016single,shi2019counting, liu2019context,liu2020crowd} assumes pixel-wise independence, which supervises the predicted density map based on the individual pixels. However, it has two significant weaknesses. Firstly, the predicted density map tends to be over-smooth \cite{liu2018decidenet}; specifically, it may underestimate high-density level regions and overestimate low-density level regions. As a result, the model may primarily focus on achieving lower count errors rather than regressing high-quality density maps; thus, it cannot reflect the actual density levels. Secondly, without a large receptive field, individual pixel-wise loss functions may ignore the regional density level information during the training process \cite{jiang2020attention}. Unbalanced low and high-level density distributions may introduce significant bias in the training process, thus weakening the network's robustness. 
To address these issues, we proposed a novel loss function for density map regression, called Dilated Contrastive Density Loss $(L_{DCD})$, where the density difference among dilated adjacent pixels is utilized to provide additional regional supervision. Ablation studies demonstrate that our proposed regional loss function can improve the counting performance of the pixel-wise loss supervised methods.

In summary, this work makes the following contributions:
1.) We addressed the feature learning issues of the backbone network of the auxiliary task in crowd counting challenges, by enabling task-shareable and task-specified feature learning simultaneously with a primary focus on the main task. 
2.) We proposed crowd segmentation and density level segmentation as auxiliary tasks in crowd counting with additional spatial crowd location and density level information enhancement. Moreover, a \textit{GCN} model was proposed to reason about the cross-domain feature relations between density map regression and other auxiliary tasks. 
3.) We proposed a novel loss function tailored for density map regression, strengthening the network's generalizability and improving the counting accuracy.
4.) We conducted extensive experiments on seven well-known challenging counting benchmarks. Quantitative and qualitative results demonstrated that our model achieves state-of-the-art performance. 
Especially, to the best of our knowledge, we achieved the best counting performance among auxiliary tasks based counting methods on NWPU-Crowd \cite{gao2020nwpu} benchmark \footnotemark\footnotetext{\url{https://www.crowdbenchmark.com/nwpucrowd.html}}, which is currently the largest crowd counting benchmark. Our model is robust and generalizable to indicate the wrong labeled or miss labeled object in the test datasets. Please refer to Section~\ref{sec:discussion} for more details.

\section{Related Work}
In recent years density map regression-based counting methods with \textit{CNN} achieved good performance. 
For example, \textit{Boominathan et al.} \cite{boominathan2016crowdnet} proposed a dual-column network to combine low-level and high-level features in different layers to estimate the count. 
However, because of the conflicts from optimization among different columns \cite{babu2018divide}, these types of network structures have difficulty in attaining global minimization.   
Other works employed single column network structures and handled different scale challenges \cite{yang2020reverse,ma2020learning,zhao2019scale,zhou2020adversarial,liu2019crowd,liu2018crowd} with adaptive modules, such as scaled spatial pyramid pooling \cite{liu2019context,tian2019padnet,chen2019scale} or Dilated kernels of filters \cite{li2018csrnet,yan2019perspective,yan2021crowd,wan2020kernel,bai2020adaptive}.
They achieved promising counting performance along with architectural simplicity and training efficiency. 

\subsection{Attention-Based Counting}
The visual attention mechanism was applied among several works \cite{miao2020shallow,jiang2020attention,zhang2019attentional,chen2021variational,duan2020sofa,wang2020density,wan2020kernel} in the crowd counting task, which helped the network focus on valuable information and addressed several challenges. For example, \textit{Miao et al.} \cite{miao2020shallow} utilized a shallow feature based attention module to highlight the regions of crowd interest and filter out the noise in the background clutter. To tackle various density levels issues, \textit{Jiang et al.} \cite{jiang2020attention} employed an attention mask to refine the density map for adapting to different density levels. Furthermore, \textit{Zhang et al.} \cite{zhang2019attentional} proposed the \textit{Attention Neural Field} that incorporated non-local attention modules and conditional random fields to maintain multi-scale features and long-range dependencies, to handle large scale changes problems of the input crowd images. \textit{Wan et al.} \cite{wan2019adaptive,wan2020kernel} exploited the self-attention mechanism to adaptively generate density maps with different Gaussian kernel sizes, which is used as the ground truth to supervise the model.
The aforementioned methods adopted the attention mechanism as a feature enhancement module to implicitly address the crowd counting task's challenges, such as significant scale changes, wide variations of density levels, and complex scene backgrounds. Our model explicitly addressed those challenges through auxiliary tasks. On the other hand, our model adopted the attention mechanism to construct an adaptively shared backbone network, enabling the task-shared and task-specific features learning simultaneously.

\subsection{Auxiliary Tasks Based Counting}
Recently, auxiliary task learning based counting methods \cite{yang2020embedding,cheng2021decoupled,abousamra2021localization,song2021choose,liu2021cross,zhang2021cross,lei2021towards,wan2021fine,jiang2020density,liu2021counting,liu2020estimating} 
attracted researchers' attention because of its ability to capture extra granularity information and contextual dependencies for the density map regression. Most of the methods utilized the potential of a model itself with auxiliary tasks, such as object detection, crowd segmentation, density level classification, \(etc.\), to enhance the feature tuning for density map regression. 
For example, the task of patch-based density level classification \cite{shi2019counting,sindagi2017generating,sindagi2019ha,jiang2020density,zhou2021locality,liu2020adaptive,mo2020background} can enhance patch-level density level information, which helped to address the underestimation and the overestimation problems of density map regressions. However, it may be difficult to guide the pixel-wise density map regression via patch-wise density level classification because of the gap between pixel-level and patch-level feature learning. In contrast, our model proposed a density level segmentation auxiliary task, which can be regarded as the dense pixel-wise density level classification task. In this way, our model can enhance the pixel-wise density level information to the pixel-wise density map regression task, aiming to address the challenges of wide variations of density levels.

Moreover, because the background regions in complex scenes contain confusing objects or similar appearance, the crowd segmentation task, adopted by previous methods \cite{zhao2019leveraging,shi2019counting,luo2020hybrid,modolo2021understanding,wang2021pixel}, can provide spatial location information for the crowd, which highlighted the foreground over the background and guided the network focus on the region of interest. Our model also adopted the crowd segmentation task because of its superiority in spatial location information enhancement.
Similarly, the task of object (crowd) detection \cite{liu2018decidenet,sam2020locate,liu2019recurrent,lian2019density,wang2021self,ren2020tracking} can enhance location information and alleviate local spatial inconsistency issues in the density map. 

\subsection{Learn to Count with Different Supervisions}
Instead of tackling the counting task through different learning frameworks or strategies, recent methods \cite{liu2019exploiting,liu2018leveraging,Sravyaacmm2021,song2021rethinking,wan2021generalized,ma2019bayesian,wan2020modeling,wang2020distribution,wang2021uniformity,ma2021spatiotemporal} payed attention at the way of supervisions. For example, 
\textit{Sravya et al.} proposed a bin loss \cite{Sravyaacmm2021} to enable the data-distribution aware optimization, which helped to address the domain variation challenges from different crowd data source.
\textit{Song et al.} \cite{song2021rethinking} studied the counting problem in a different way, where a combination of \textit{Euclidean} loss and \textit{Cross Entropy} loss was used for point locations learning, instead of density map regression.
Along the same line, \textit{Bayesian} loss was proposed by \cite{ma2019bayesian} to provide more reliable supervisions at each annotated point. Differently, \textit{Wan et al.} \cite{wan2021generalized} studied the combination of pixel-wise loss and point-wise loss, which investigated the density map representation through an unbalanced optimal transport problem. \cite{wan2020modeling} proposed a novel loss function to address the spatial annotation noise during training, where a weighted MSE term and a pixel-wise correlation term were involved. Recently, \cite{wang2020distribution} proposed distribution matching loss to tackle the weakened generalizability of the Gaussian smoothed density map.
Moreover, \textit{Wang et al.} \cite{wang2021uniformity} treated the counting with density map as a classification problem, where a Cross-Entropy loss was used to classify each patch into certain intervals.

The aforementioned methods introduced different loss functions to supervise the model, such as points location, bounding box, matching, ranking, classification, \textit{etc.}. However, the mainstream counting methods still relies on pixel-wise supervision with the density map ground truth \cite{gao2020cnn}, such as \textit{L1} or \textit{L2} loss functions. In this work, we proposed a Dilated Contrastive Density Loss ($L_{DCD}$) to improve the pixel-wise loss's receptive field and increase the regional supervision. 

\begin{figure*}[t]
\centering
\includegraphics[width=18cm]{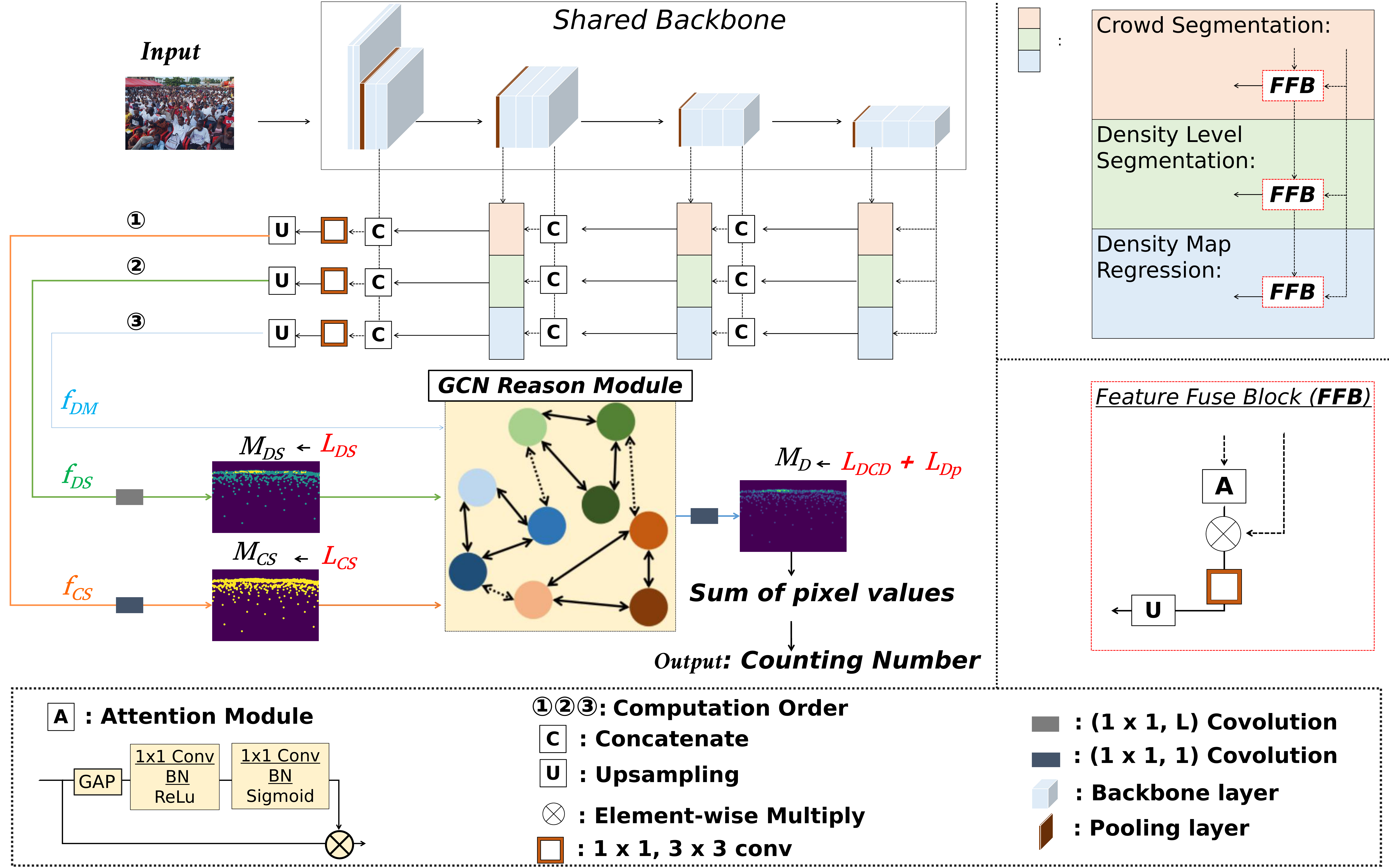}
\caption{Illustration of our proposed network. The adaptively shared backbone network has three outputs of \(f_{CS}, f_{DS}, f_{DM}\), representing crowd segmentation, density level segmentation, and density map regression branches' output feature map, respectively. The order of their involvements indicates that the density map regression branch can benefit from the extra density level and crowd spatial supervision from the other two branches gradually. }
\label{networkstructure}
\end{figure*}
\section{Methodology}

\subsection{Ground Truth Generation}
Following \cite{lempitsky2010learning}, given a set of \(N\) images \(\{I_i\}_{i=1}^{N}\) with corresponding point annotations \(\{P_i\}_{i=1}^{N}\), the ground truth of the density map \(\{D_i\}_{i=1}^{N}\) is generated by filtering the points with a normalized Gaussian kernel. The total object count number \(T_i\) of image \(I_i\) can be attained by summing all pixel values of the density map \(D_i\). 

The ground truth mask of the crowd segmentation task is generated from the density map ground truth. Given a set of \(N\) density maps \(\{D_i\}_{i=1}^{N}\), the value for each pixel in the mask \(\{B_i\}_{i=1}^{N}\) is set to 1 if the pixel valve in the density map is larger than the zero and 0 otherwise. 

The ground truth mask used by the density level segmentation task is also generated from the density map. For pixel \(p\) in input image \(i\), its density level class \(S_{p,i}\) is given as:
\begin{equation}
    S_{p,i} =\min\limits_{i = 1,..,N}{\left(\floor{\frac{D_i(p) - \min(D_i)}{ \max(D_i) - \min(D_i)} \times L + 1}, L + 1\right)},
\end{equation}

where \(L\) represents the overall levels of density; following previous patch-based density level classification methods \cite{shi2019counting,jiang2020density}, we set \(L\) equal to 4 in our work. \(D_i\) is the pixel value in the \(i_{th}\) density map ground truth.

\subsection{Task Adaptive Backbone Network}
Instead of using a shared backbone network to extract generalizable features for different tasks, we proposed an attention-based task-adaptive shared backbone network to allow the model to extract discriminative features for the auxiliary tasks, thus helping to improve the performance of the main task.
Fig. \ref{networkstructure} shows the detailed structure of the proposed network,
which consists of a shared backbone and three attention-based task-adaptive branches. 
To make a fair comparison with previous auxiliary task-based methods, such as \cite{sindagi2017generating,luo2020hybrid,sam2020locate,jiang2020density}, \(etc.\), the truncated VGG-16 \cite{Simonyan15} is used as the backbone network. However, it can be replaced by any other robust network structure; we have reported the counting performance with other powerful network backbones in TABLE. \ref{morebackbone}. The shared backbone adopts the first 13 layers of the VGG-16 to extract multi-level features. To exploit the global contextual dependencies, we proposed a Feature Fuse Block (\textit{FFB}), which aggregated and fused the outputs from posterior layers back to the preceding layers hierarchically and iteratively, with up-sampling, concatenation and convolution operations. 
This provides improvements in extracting the full spectrum of semantic and spatial information across different stages and resolutions. 
The up-sampling is performed by using a bilinear interpolation algorithm. The convolution operation aims to reduce and match the corresponding feature map channel size between different stages.

With the aggregating process from low-level features to high-level features, the task-adaptive attention module is applied in three different task branches; details of the attention module are shown in the bottom left of Fig. \ref{networkstructure}. Each attention module consists of a global average pooling (GAP) layer to capture global context through different feature map channels, conducting an attention tensor to lead the emphasis of feature learning. Then, two blocks with a convolutional layer followed by a Batch Normalization \textit{(BN)} \cite{pmlr-v37-ioffe15} layer with \textit{ReLu} and sigmoid as the activation functions are added. For the convolutional layer filter, the kernel size is \(1 \times 1\). The element-wise multiplication is then performed between the outputs of the particular layer of the shared backbone and the task-specific attention module, which filters out the unrelated and redundant features from the backbone with respect to different auxiliary tasks and the main task. Therefore, the shared backbone can learn a generalizable representation, while the attention-based branches can extract task-specific features simultaneously in an end-to-end manner. The ablation study experiments proved that the attention-based adaptive backbone could boost the counting performance. 

Apart from the aforementioned network structure component in three attention-based task-adaptive branches, we also introduced a cross-domain feature fusing operator in a particular order to focus on optimizing the primary density map regression task primarily. Specifically, the crowd segmentation branch is applied to the shared backbone first to select the corresponding discriminative spatial features. Then, we applied the density level segmentation branch on the shared backbone and crowd segmentation branch, which can enhance the additional contextual density level information into the main task. At last, the main task of the density map regression branch is applied. 



\subsection{Auxiliary Tasks}
With three outputs from the task adaptive backbone network, we built two auxiliary tasks and a main task: crowd segmentation task, density level segmentation task, and density map regression task. We detail each of them subsequently.

\noindent\textbf{Crowd Segmentation.} We introduced crowd segmentation as one of the auxiliary tasks for two reasons. Firstly, the density map's pixel value should be zero at non-crowd regions. However, the predicted density map can be noisy and inaccurate when the background is cluttered and complex. The crowd segmentation task provides a spatial focus to the density map regression process through zero out the non-crowd regions' pixel values. 
Secondly, given the standard set-up of single density map regression, the pixels within a particular range of the point annotations should contribute more to the final counting results; however, the loss is dominated by the majority of less relevant pixels. To overcome this limitation, the crowd segmentation can provide additional information enhancement in terms of the spatial indicator with a standalone loss function.

Given an input image $I_{i}\in{\mathbb{R}^{3\times{H}\times{W}}}$ , we can get the output of the crowd segmentation branch in the backbone network, \(f_{CS}\in{\mathbb{R}^{C\times{H}\times{W}}}\), where $H$ and $W$ represent the height and width of the feature map; \textit{C} is the channel size. Then, we apply a convolution layer with filter parameters \(\theta_{CS}\in{\mathbb{R}^{1\times{1}\times{1}}}\), followed by a sigmoid as the activation function. Through this operation, we can generate a probability map to calculate the crowd and background probability. The single channel crowd segmentation probability map \(M_{CS}\) is defined as: \(M_{CS} = Sigmoid(\theta_{CS},f_{CS})\in{\mathbb{R}^{1\times{H}\times{W}}}\). 

\noindent\textbf{Density Level Segmentation.} Density map regression is a pixel-wise task, which focuses on low-level features learning but may ignore high-level contextual information during the training \cite{tian2019padnet}. To address this issue, we perform density level segmentation as another auxiliary task. Compared with previous patch-based density level classification methods \cite{shi2019counting,sindagi2017generating,sindagi2019ha,jiang2020density}, our proposed pixel-based density level segmentation can provide pixel-wise level density information and high-level semantic features at the same time.
Upon the output of the density level segmentation branch of the backbone network \(f_{DS}\in{\mathbb{R}^{C\times{H}\times{W}}}\), a convolution layer with filter parameters \(\theta_{DS}\in{\mathbb{R}^{ L\times1\time1\times1}}\) and a softmax as activation function are applied. The prediction of density level segmentation branch \(M_{DS}\) is defined as: \(M_{DS} = softmax(\theta_{DS},f_{DS}) \in{\mathbb{R}^{L\times{H}\times{W}}}\), where $L$ is the number of density levels.

\begin{figure}
\centering
\includegraphics[width=8cm]{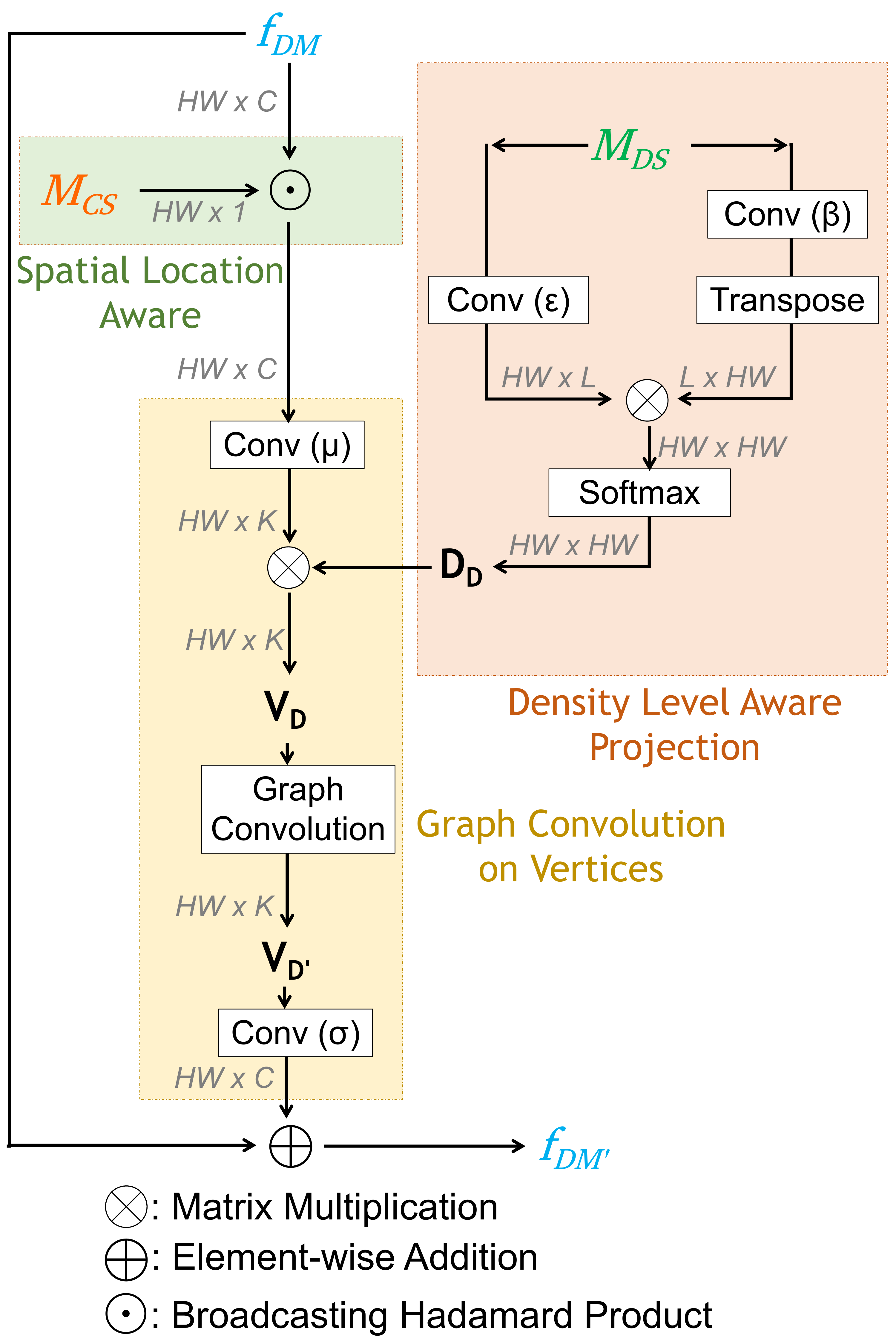} 
\caption{Architecture of the proposed \textit{GCN} reasoning module. \(f_{DM}\in{\mathbb{R}^{C\times{H}\times{W}}}\) is the feature map of the density map regression branch, \(C = 32\) is the channel size; \(M_{CS}\in{\mathbb{R}^{1\times{H}\times{W}}}\) is the prediction of the crowd segmentation branch; \(M_{DS}\in{\mathbb{R}^{L\times{H}\times{W}}}\) is the prediction of density level segmentation branch, \(L = 4\) is the number of density levels; \(D_{D}\in{\mathbb{R}^{HW\times{HW}}}\) is the density level dependency matrix; \(V_{D}\in{\mathbb{R}^{K\times{HW}}}\) is the constructed vertex features and \(V_{D'}\in{\mathbb{R}^{K\times{HW}}}\) is the output vertex features after \textit{GCN}, \(K = 16\) is the number of vertices. \(f_{DM'}\in{\mathbb{R}^{C\times{H}\times{W}}}\) is the output feature map after \textit{GCN} reasoning.}
\label{gcn_module}
\end{figure}

\subsection{Density Map Regression} 
Intuitively, the different granularity features of density levels and spatial crowd locations need to be further reasoned to fuse into the density map regression branch. To this end, with the predicted crowd segmentation output $M_{CS}$ and density level segmentation output $M_{DS}$ as the auxiliary information granularity, we input them along with the density map branch's feature map \(f_{DM}\in{\mathbb{R}^{C\times{H}\times{W}}}\) into the \textit{GCN} reasoning module to reason the relationship among themselves. Subsequently, the output feature map \(f_{DM'}\in{\mathbb{R}^{C\times{H}\times{W}}}\) of the \textit{GCN} reasoning module is reduced into one-channel through \(1\times1\) convolution layer with a \textit{ReLU} activation function.

\subsection{\textit{GCN} Reasoning Module}
The proposed \textit{GCN} reasoning module structure is shown in Fig. \ref{gcn_module}. In detail, there are three primary modules: \textit{Spatial Location Aware} module, \textit{Density Level Aware Projection} module, \textit{Graph Convolution on Vertices} module. 
Because of the nature of the crowd images, the density level varies across the image \cite{liu2019context}, which indicates that the pixel values of the density map should not just rely on their own pixel-wise features but also on different density level regions. To this end, our \textit{GCN} reason model projected a collection of spatial-aware density feature map's pixels with similar density levels to each graph vertex and exploited a \textit{GCN} to reason about the relations among graph vertices. 

\noindent\textbf{Spatial Location Aware Module.} Before projecting the density map feature map \(f_{DM}\) into the graph vertices, we directly applied the broadcasting Hadamard Product operation between the crowd segmentation output \(M_{CS}\) and the density map regression branch's feature map \(f_{DM}\).
There are two underlying reasons: (1) \(M_{CS}\) is a one-channel crowd segmentation map, with encoding the probability of the non-crowd regions' pixel values approaching zero and crowd regions' pixel values approaching one; one can serve as a filter to zero out the non-crowd region's pixel value of the density map. (2) Direct broadcasting Hadamard Product can achieve crowd spatial awareness for every channel of the \(f_{DM}\) through zero out the non-crowd region's pixel value. This can address the challenge of the complex scene backgrounds in the crowd images.

\noindent\textbf{Density Level Aware Projection Module.} As mentioned above, the pixel-wise density level information can help to address the challenges of the large variations of density levels in crowd images. However, direct broadcasting Hadamard product between the density map branch's feature map \(f_{DM}\) and the density level output \(M_{DS}\) may result in domain conflicts \cite{luo2020hybrid}. We exploited the nature of \textit{GCN} and projected the density level information into the graph vertices for further reasoning; one benefits the long-range relationship reasoning ability of \textit{GCN} and the multi-granularity information enhancement from density level. Inspired by the non-local module \cite{wang2018non}, we encoded the long-range density level dependency among every pixel. Give the feature map \(M_{DS}\), the density level dependency matrix \(D_{D}\in{\mathbb{R}^{HW\times{HW}}}\) is defined as: 
\begin{equation}
    D_{D} = softmax\Big(\epsilon(M_{DS}) \otimes \beta^\mathsf{T}(M_{DS})\Big),
\end{equation}
where Conv \(\beta\) and Conv \(\epsilon\) are two convolution layers with \(1 \times 1\) kernel size, respectively. The dependency matrix \(D_{D}\)
can be regarded as a pixel-wise attention map, where pixels with similar density levels are assigned with larger weights. The dependency matrix itself can reflect the pixel-wise density level dependency. Besides, we projected it as a prior to the graph domain through matrix multiplication, which can enhance high-level contextual dependency simultaneously. 

\noindent\textbf{Graph Convolution on Vertices.}
In this module, we learnt how to reason the region-based relationship in density map through \textit{GCN} in graph domain. Firstly, we projected the spatial aware feature map of \(f_{DM}\) into graph domain with \(K\) vertices, and each vertex was represented by an embedding of shape \(H \times W\). This is achieved by Conv (\(\mu\)), which is a \(1\times1\) convolution layer. Furthermore, we projected the dependency matrix \(D_{D}\) to the graph domain through matrix multiplication, resulting in the vertex features \(V_{D}\in{\mathbb{R}^{K\times{HW}}}\). The projection aggregated pixels with similar density levels to graph vertices, where each vertex represents a region in the crowd image. Formally, \(V_{D}\) is defined as:
\begin{equation}
    V_{D} = D_{D}\otimes \mu(f_{DM} \odot M_{CS}),
\end{equation}
where \(\otimes\) is matrix multiplication; \(\odot\) is broadcasting Hadamard product.
With the constructed vertices, the long-range region-wise relationship is further reasoned in the graph domain through \textit{GCN}. In detail, we reasoned over the region-wise relations by propagating information across vertices with a single layer \textit{GCN}. Specifically, we fed the constructed vertex features \(V_{D}\) into a first-order approximation of spectral graph convolution \cite{kipf2016semi}, resulting the output vertex features \(V_{D'}\in{\mathbb{R}^{K\times{HW}}}\). The \(V_{D'}\) is calculated as:
\begin{equation}
    V_{D'} = ReLU\Big((I - A)\otimes{V_{D}}\otimes{W_{D}}\Big),
\end{equation}
where \(I\) is the identity matrix; \(A\in{\mathbb{R}^{HW\times{HW}}}\) denotes the adjacent matrix that encodes the graph connectivity to learn; \(W_{D}\in{\mathbb{R}^{K\times{K}}}\) is the weights of the \textit{GCN}. The adjacent matrix \(A\) is randomly initialized but can learn and update the edge weights from vertex features by gradient along the training process. The identity matrix \(I\) serves as a residual connection that alleviates the optimization difficulties. Based on the learned graph, the information propagation across all vertices leads to the finally reasoned relations between regions. After graph reasoning, a collection of pixels embedded within one vertex share the same context of features modeled by graph convolution. Then, we re-projected the vertex features in the graph domain to the original pixel grids. Given the reasoned vertices \(V_{D'}\), we applied Conv (\(\sigma\)), which is a \(1 \times 1 \) convolution layer. 
Finally, we summed up the re-projected refined and the original density feature maps as the final feature map. The final pixel-wise density feature map \(f_{DM'}\) is thus computed by
\begin{equation}
    f_{DM'} = f_{DM} + \sigma (V_{D'}).
\end{equation}

\begin{figure}
\centering
\includegraphics[width=9cm]{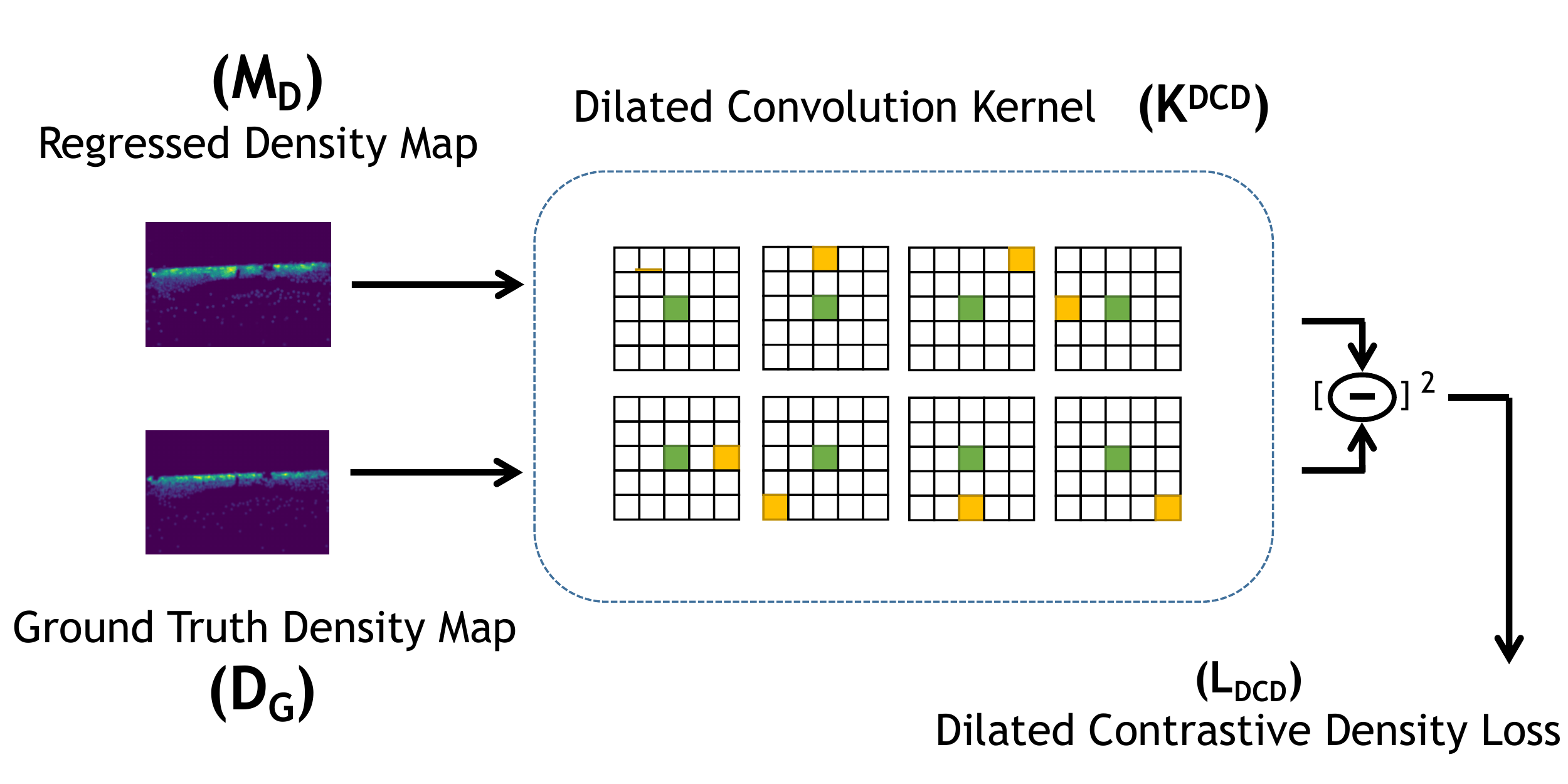} 
\caption{Dilated Contrastive Density Loss (\(L_{DCD}\)). There are eight dilated contrastive kernels with green, white, yellow blocks representing 1, 0, -1, respectively. The least-square error of two outputs from regression and ground truth is treated as the final \(L_{DCD}\).}
\label{contrastive_loss}
\end{figure}
\subsection{Loss Function}
The whole network is end-to-end trainable, which includes four loss functions; the total loss function is defined as:
\begin{equation}
    L_{total} =  L_{CS} +  L_{DS} + \gamma \cdot ( L_{Dp} + L_{DCD}),
\end{equation}
where \(\gamma\) is empirically set as 2, which is a hyper-parameter to trade-off between auxiliary losses and main loss. Please note that, extensive experiments have been done to determine the weights of the losses for two auxiliary tasks, respectively. We found that there is no significant difference of counting performance with respect to different weight values; thus, we set them both equal to 1 in the loss function.  Binary cross-entropy \((L_{CS})\) is used for crowd segmentation auxiliary task; categorical cross-entropy \((L_{DS})\) is used for density level segmentation auxiliary task; \textit{L2} loss is used for pixel-wise density map regression supervision (\(L_{Dp}\)). However, pixel-wise \textit{L2} loss assumes pixel-wise independence, which results in over-smooth density map prediction \cite{liu2018decidenet} and the underlying bias from unbalanced low- and high-level density distributions of crowd images. 
To address the issue, we proposed Dilated Contrastive Density Loss \((L_{DCD})\), where we take into account more adjacent pixels for regional density difference. In detail, we applied single layer convolution on the regressed density map \(M_{D}\) and the ground truth density map \(D_{G}\) respectively. The single layer convolution has eight filters; each filter contains dilated kernel with a fixed value (\(e.g.\) 1, 0, and -1). The least-square error of the calculated regional dilated contrastive values from the regressed and ground truth density map is the output of \(L_{DCD}\). To this end, we define
\(L_{DCD}\) as below:
\begin{equation}
    L_{DCD} = \sum_{i}\vert \vert K_{i}^{DCD} \otimes M_{D} - K_{i}^{DCD} \otimes D_{G} \vert \vert _{2}^{2},
\end{equation}
where \(K_{i}^{DCD}\) is the \(i^{th}\) dilated contrastive convolution kernel, \(i \in{[1,8]}\). Details of the kernel are shown in Fig. \ref{contrastive_loss}, where a \(3 \times 3\) convolution layer with the dilated rate of \(2\) is applied; one gives a larger receptive field as \(5 \times 5\). We perform extensive experiments to evaluate the effectiveness of the proposed \(L_{DCD}\) loss; quantitative results in \textit{Ablation Study} (Section~\ref{sec:ablation}) demonstrates that the proposed \(L_{DCD}\) loss can improve the counting accuracy not only for our model but also for previous single \textit{L2} loss based methods.

\begin{figure*}
\centering
\includegraphics[width=17cm]{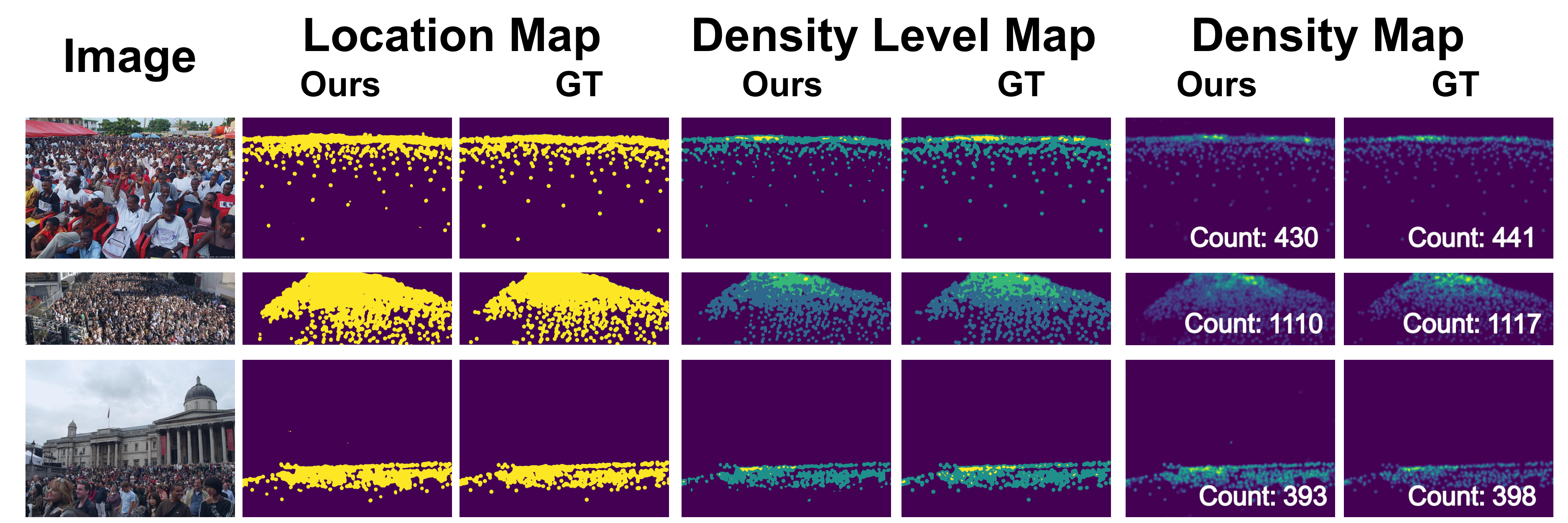}
\caption{Qualitative results of the density, crowd location and density level map in \textit{SHA} test dataset. Our model can produce accurate density maps compared to the ground truth (\textit{GT}), along with accurate auxiliary crowd segmentation and density level segmentation results. 
}
\label{allresults}
\end{figure*}

\begin{table*}[th]
\centering
\caption{Results on five challenging datasets for crowd counting, compared with other \textbf{auxiliary task learning based methods}. Our model achieves a new state-of-the-art within auxiliary learning based counting methods in terms of \textit{MAE}.}
\scalebox{1.3}
{
\begin{tabular}{ccc cc cc cc cc}
\hline
\multirow{2}{*}{Methods}  & \multicolumn{2}{c}{\textit{SHA}} & \multicolumn{2}{c}{\textit{SHB}} & \multicolumn{2}{c}{\textit{QNRF}} & \multicolumn{2}{c}{\textit{JHU-Crowd}} & \multicolumn{2}{c}{\textit{NWPU-Crowd}} \\ \cline{2-11} 
&\multicolumn{1}{c}{\textit{MAE}} & \textit{RMSE}  & \multicolumn{1}{c}{\textit{MAE}} & \textit{RMSE} & \multicolumn{1}{c}{\textit{MAE}} & \textit{RMSE} & \multicolumn{1}{c}{\textit{MAE}} & \textit{RMSE} & \multicolumn{1}{c}{\textit{MAE}} & \textit{RMSE} \\ \hline
\textit{CP-CNN} \cite{sindagi2017generating} & 73.6 & 106.4 & 20.1 & 30.1 & - & - & - & - &- &- \\
\textit{DecideNet} \cite{liu2018decidenet}   & - & - & 21.53 & 31.98 & - & - & - & - &- &-\\
\textit{CFF} \cite{shi2019counting}   & 65.2 & 109.4 & 7.2 & 12.2 & 93.8 & 146.5 & 83.6 & 400.7 &80.8 &364.1\\
\textit{AT-CSRNet} \cite{zhao2019leveraging}  & - & - & 8.11 & 13.53 & - & - & - & - &- &-\\
\textit{SHRGBD} \cite{lian2019density}  & 70.3 & 111.0 & 8.8 & 15.3 & 113.3 & 177.6 & 107.9 & 446.7 & 103.0 & 478.1\\
\textit{HA-CCN} \cite{sindagi2019ha}  & 62.9 & 94.9 & 8.1 & 12.7 & 118.1 & 180.4 & - & - &- &-\\
\textit{RAZ-Net} \cite{liu2019recurrent}  & 65.1 & 106.7 & 8.4 & 14.1 & 116 & 195 & - & - & 151.5 &634.6\\
\textit{HYGNN} \cite{luo2020hybrid}  & 60.2 & 94.5 & 7.5 & 12.7 & 100.8 & 185.3 & - & - &- &-\\
\textit{LSC-CNN} \cite{sam2020locate} & 66.4 & 117.0 & 8.1 & 12.7 & 120.5 & 218.2 & 112.7 & 454.4 &90.4 &388.8 \\
\textit{ASCC} \cite{jiang2020attention} &57.8 & \textbf{90.1} &7.5 &13.1  &91.6 & 159.7 &84.6 &355.1 &95.7 &398.0\\
\textit{UMRNet} \cite{modolo2021understanding}  & 62.6 &103.3 &7.2 & 11.5 & 86.3 &153.1 & - & - & - & -\\
\textit{DAMNet} \cite{jiang2020density} & 63.1 & 106.3 & 9.1 & 16.3 & 101.5 & 186.9 & - & - & - &- \\
\textit{MATT} \cite{lei2021towards} & 59.5 & 97.3 & \textbf{6.9} & \textbf{10.3} &- & - &- & - & -&-
\\ \hline \hline
\textit{Ours} & \textbf{57.0} & 98.6 & 7.1 & 12.3  & \textbf{85.3} & \textbf{129.4} & \textbf{66.6} & \textbf{254.9}  & \textbf{76.4}  & \textbf{327.1}  \\ \hline
\end{tabular}
}
\label{totalresult}
\end{table*}

\begin{table*}[th]
\centering
\caption{Results of using different backbone networks on five crowd counting datasets. }
\scalebox{1.3}
{
\begin{tabular}{c||cc|cc|cc|cc|cc}
\hline
\multirow{2}{*}{Methods}  & \multicolumn{2}{c|}{\textit{SHA}} & \multicolumn{2}{c|}{\textit{SHB}} & \multicolumn{2}{c|}{\textit{QNRF}} & \multicolumn{2}{c|}{\textit{JHU-Crowd}} & \multicolumn{2}{c}{\textit{NWPU-Crowd}} \\ \cline{2-11} 
&\multicolumn{1}{c|}{\textit{MAE}} & \textit{RMSE}  & \multicolumn{1}{c|}{\textit{MAE}} & \textit{RMSE} & \multicolumn{1}{c|}{\textit{MAE}} & \textit{RMSE} & \multicolumn{1}{c|}{\textit{MAE}} & \textit{RMSE} & \multicolumn{1}{c|}{\textit{MAE}} & \textit{RMSE} \\ \hline
\textit{VGG-16} \cite{Simonyan15} & \textbf{57.0} & 98.6 & 7.1 & 12.3  & \textbf{85.3} & 129.4 & \textbf{66.6} & 254.9  & \textbf{76.4}  & \textbf{327.4} \\
\textit{VGG-19} \cite{Simonyan15} & 59.7 & 99.8 & 8.4 & 13.2 & 87.8 & 144.0 & 73.7 & 320.1 &79.9 &360.0\\
\textit{ResNet-50} \cite{he2016deep} &57.8 &\textbf{96.6} &\textbf{7.0} &\textbf{11.7} &85.5 &\textbf{128.7} &77.9 &318.1 &79.3 &344.4 \\
\textit{ResNet-101} \cite{he2016deep} &61.1 &100.8 &9.1 &14.5 &93.3 &147.9 &69.7 &\textbf{253.3} &81.4 &361.5 \\ \hline
\end{tabular}
}
\label{morebackbone}
\end{table*}

\section{Experiments}
\subsection{Datasets}
\textbf{ShanghaiTech} \cite{zhang2016single} consists of 1,198 images, containing a total amount of 330,165 people with head centre point annotations. This dataset is divided into two parts: \textbf{\textit{SHA}} includes 482 images, in which crowds are mostly dense (33 to 3139 people); \textbf{\textit{SHB}} includes 716 images, where crowds are sparser (9 to 578 people). Each part is divided into training and testing subset as specified in \cite{zhang2016single}. \textbf{\textit{UCF-QNRF}} \cite{idrees2018composition} is a large crowd dataset, consisting of 1,535 images with about 1.25 million annotations in total. The number of people in these images varies largely with a wide range from 49 to 12,865. As indicated by \cite{idrees2018composition}, For training, 1,201 images are used, the remaining 334 images form the test set. \textbf{\textit{JHU-Crowd}} \cite{sindagi2019pushing} is a recent challenging large-scale dataset that containing 4,372 images with 1.51 million annotations. The dataset includes several challenging scenes such as weather-based degradation and illumination variations \(etc.\). This dataset is divided into 2,272 images for training, 500 images for validation, and 1,600 images for testing. \textbf{\textit{NWPU-Crowd}} \cite{gao2020nwpu} is up to date the largest public crowd counting dataset, containing 5,109 images with over 2.13 million annotations. The dataset includes 3,109 training images, 500 validation images and 1,500 test images. 
Moreover, inspired by the potential of crowd counting, we conducted experiments on commonly used cell counting dataset: 
\textbf{\textit{DCC}} \cite{marsden2018people} with 100 images for training and 77 images for testing, and vehicle counting dataset: \textbf{\textit{Trancos}} \cite{guerrero2015extremely} with 403 images for training, 420 images for validation and 421 images for testing. These experiments further demonstrated our model's robustness and applicability for different real-world applications. 

Note that, for ShanghaiTech (\textit{SHA}, \textit{SHB}), \textit{UCF-QNRF}, and \textit{DCC} dataset, we use 10\% of the given training images as the validation dataset. 

\subsection{Implementation Details} \label{sec:implementation}
To augment the dataset, we randomly cropped the input images, density maps, crowd segmentation masks, and density level segmentation masks with fixed size \(128 \times 128\) at a random location, then randomly horizontally flipped the image patches with the probability of 0.3. 
We trained our model for 400 epochs for all experiments, with a start learning rate of $1e-4$ and a cosine decay schedule \cite{loshchilov2016sgdr}. The batch size is set to 96. All the training processes are performed on a server with 8 TESLA V100 and 4 TESLA P100, and all the test experiments are conducted on a local workstation with a Geforce RTX 2080Ti. Five-fold cross-validation is used for fair comparison and hyper-parameters tuning in all settings.

\subsection{Evaluation Metrics}
To evaluate the counting performance, we adopted Mean Absolute Error \((MAE)\) and Root Mean Squared Error \((RMSE)\). 
Since Mean Absolute Error (\(MAE\)) and Root Mean Square Error (\(RMSE\)) cannot measure the counted objects’ locations, Grid Average Mean absolute Error (\(GAME\)) is used to indicate counting accuracy over local regions. 
\(GAME\) is defined as:
\begin{equation}
    GAME(L) = \frac{1}{N} \sum_{n=1}^{N}(\sum_{l=1}^{4^{L}}|y^{l}_{n} - \hat{y}^{l}_{n}|),
\end{equation}
where N is the total number of images, \(y_{n}^{l}\) and \(\hat{y}_{n}^{l}\) are the ground truth and estimated counts in the local region \(l\) of \(n^{th}\) image. \(4^{L}\) denotes the number of non-overlapping regions which cover the full image. When L equals to 0, the \(GAME\) is equivalent to \(MAE\).

\begin{figure}[ht]
\centering
\includegraphics[width=9cm]{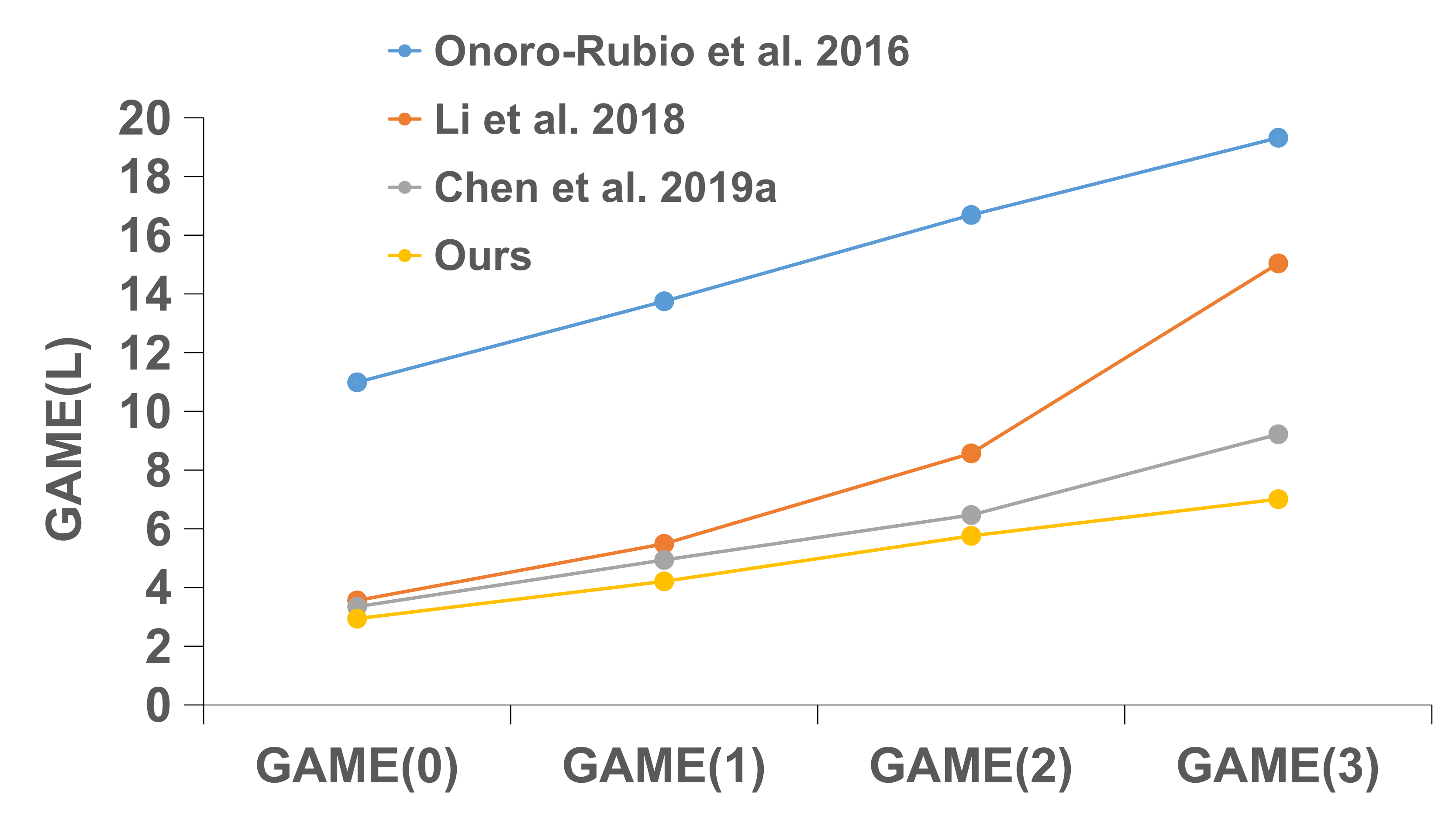} 
\caption{Comparison of \(GAME\) performance on the \textit{Trancos} dataset among the proposed approach and the state-of-the-arts, such as \textit{Onoro-Rubio et al.} \cite{onoro2016towards}, \textit{Li et al.} \cite{li2018csrnet}, \textit{Chen et al.} \cite{chen2019scale}. Note that, a small range of increase among different \(GAME\) values indicates that our method counts and localizes overlapping vehicles more accurately.}
\label{GAMEresult}
\end{figure}

\section{Results}
\subsection{Counting Results}
In this section, we present our experimental results on the crowd, cell, and vehicle counting tasks in comparison to other \textbf{auxiliary-task based} state-of-the-art crowd counting methods. These experiments further demonstrate our model's robustness and applicability in multiple domain datasets.
In the Discussion (Section~\ref{sec:discussion}), we showed that our model could indicate some mislabeled or wrongly labeled point annotations from the ground truth of the test dataset. This highlights our approach's generalizability and the potential issue of imperfect ground truth in object counting datasets.%

\noindent\textbf{Crowd Counting Results.}
We performed experiments to validate our model's performance in five challenging crowd counting datasets. Fig. \ref{allresults} shows qualitative results; specifically, we presented the predictions from auxiliary task branches (crowd segmentation and density level segmentation masks) to demonstrate our model's cohesion, along with the spatial location and density level variation's contribution of auxiliary branches. To make a fair comparison, we only compared our model with previous \textbf{auxiliary task learning} based counting methods. TABLE. \ref{totalresult} shows that our method outperforms other methods in terms of \textit{MAE} on all five datasets. In particular, our model outperforms the patch-based density level classification based method \textit{HA-CCN} \cite{sindagi2019ha} by 14.7\% via average \textit{MAE}.
Notably, the \textit{JHU-Crowd} dataset \cite{sindagi2019pushing} and \textit{NWPU-Crowd} dataset \cite{gao2020nwpu} are recent public available datasets, which are more challenging due to large variations in scale, occlusion, and complex weather scenes. Specifically, \textit{NWPU-Crowd} is current the largest crowd counting benchmark \footnotemark\footnotetext{\url{https://www.crowdbenchmark.com/nwpucrowd.html}}. To the best of our knowledge, we achieved the best performance among other auxiliary task based methods.  Except the auxiliary based methods shown in TABLE. \ref{totalresult}, our method gains a superior reduction than single-task learning based methods as well, for example, scale-variation enhanced method CACC (100.1 \textit{MAE}) \cite{liu2019context} by 18.3\% and dilated kernel-based method CSR-Net (85.9 \textit{MAE}) \cite{li2018csrnet} by 4.8\% via \textit{MAE}. 

\begin{figure}
\centering
\includegraphics[width=8cm]{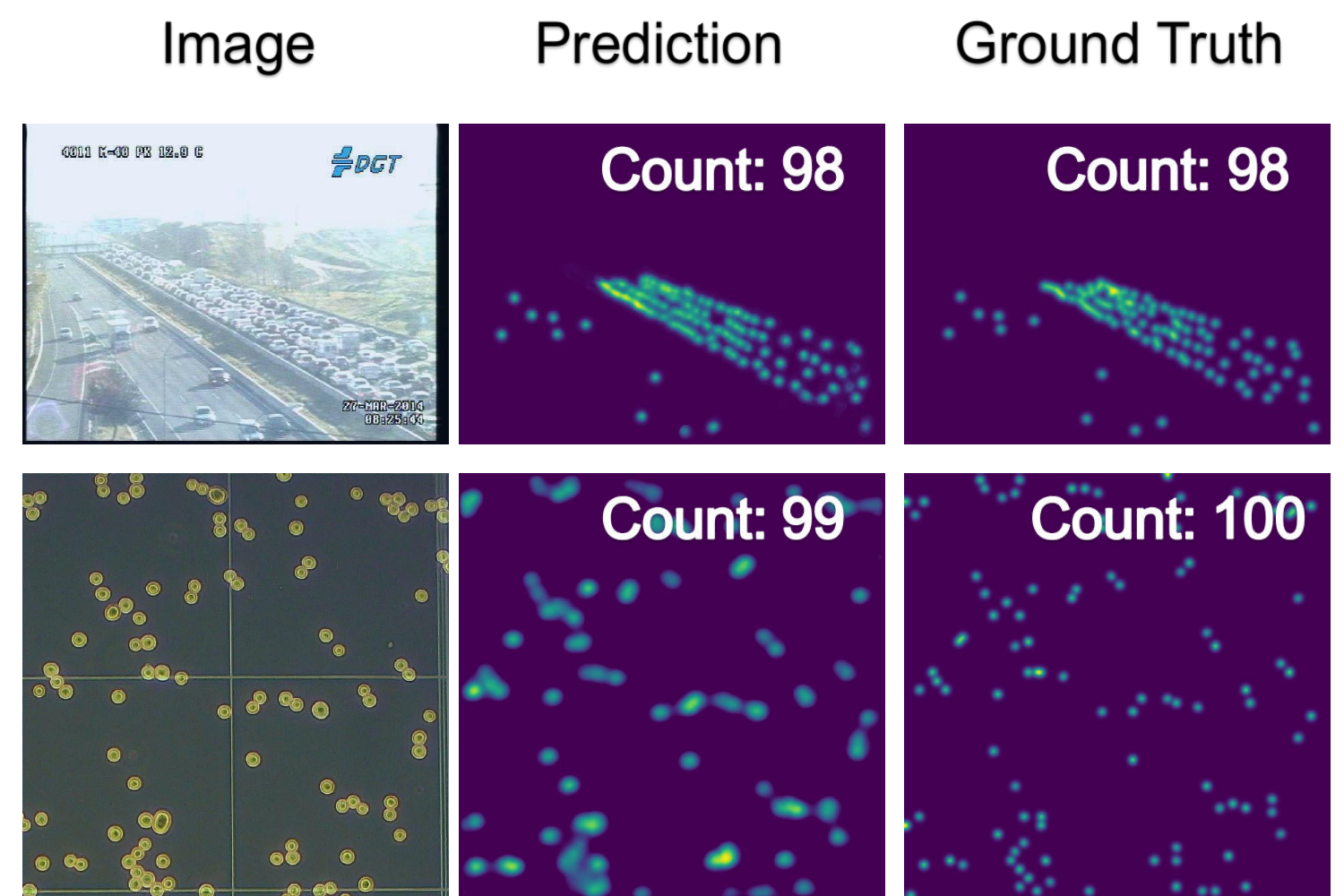}
\caption{Qualitative results on the Trancos (top) and DCC (bottom) dataset. Our model adapts well with scale variations and weather degradation challenges on the Trancos dataset. Further improvement is required for the cell dataset where individual cell locations are hard to distinguish, though the density levels and spatial distributions are clearly indicated.}
\label{cell_vehicle}
\end{figure}

\begin{table}
\centering
\caption{Results on cell (\textit{DCC}) counting and vehicle (\textit{Trancos}) counting dataset. Our model achieves superior performance to the previous state-of-the-art methods.}
\scalebox{1.2}{
\begin{tabular}{c|c|cc}
\hline
\multirow{2}{*}{Methods} & \multicolumn{1}{c|} {DCC} & \multicolumn{2}{c}{Trancos} \\ \cline{2-4} 
 & \textit{MAE} & \textit{MAE} & \textit{RMSE} \\ \hline
PPPD \cite{marsden2018people} & 8.4  & 9.7 & - \\
SAU-Net \cite{guo2019sau} & 3.0  & - & - \\
CSRNet \cite{li2018csrnet} & -  & 3.5 & 5.1 \\ 
CCF \cite{shi2019counting} & 3.2  & \textbf{2.0} & -  \\\hline
Ours & \textbf{2.9}    & 2.3 & \textbf{4.8}\\ \hline
\end{tabular}
}
\label{cellvehicle}
\end{table}

\noindent\textbf{Cell \& Vehicle Counting Results.}
We conducted experiments on cell (\textit{DCC} \cite{marsden2018people}) and vehicle (\textit{Trancos} \cite{guerrero2015extremely}) counting datasets to show our model's broad applicability and robustness. Fig. \ref{cell_vehicle} shows the qualitative results, and  Fig. \ref{cellvehicle} shows the quantitative results compared with the previous state-of-the-art methods. Due to the different scenes in the cell counting dataset, such as less occlusion, no scale variation, no complex background \(etc.\), the contribution of some components of our model will be lessened because we design our model especially for crowd counting tasks; still, our model achieves comparable performance with previous methods.
Furthermore, we presented local comparison performance through the \(GAME\) metric to indicate the model's ability to recognize the objects' locations. Fig. \ref{GAMEresult} shows the comparison results in terms of the \(GAME\) on the Trancos dataset. As illustrated, our method localizes and counts overlapping vehicles more accurately.

\subsection{Auxiliary Task Results}
In this section, we reported the performance of two auxiliary tasks. The commonly used segmentation metric Intersection over Union (\textit{IoU}) is used to evaluate the auxiliary tasks' performance. In detail, we achieved average 88.7 \% \textit{IoU} for the crowd segmentation task and 81.0 \% \textit{IoU} for the density level segmentation task on the five crowd counting datasets. Fig. \ref{allresults} shows examples of those tasks' predictions from our model.   

\subsection{Ablation Study}\label{sec:ablation}
We investigated the effect of each component in our proposed model. All ablation experiments were performed with the same settings detailed in the Implementation Details (Section~\ref{sec:implementation}). 

\noindent\textbf{Ablation on Different Network Backbones}
We evaluated the effectiveness of different backbone networks on the five crowd counting datasets. The counting performance is shown in TABLE. \ref{morebackbone} with several different backbone networks. In general, \textit{VGG}-based backbone networks achieved comparable counting performance, compared with the one of \textit{ResNet}-based backbone networks in relatively large-scale datasets, such as \textit{QNRF}, \textit{JHU-Crowd} and \textit{NWPU-Crowd}. While, \textit{ResNet}-based backbone works better on small-scale counting datasets, such as \textit{SHA} and \textit{SHB}. We reported our model's performance with \textit{VGG}-16 backbone network in TABLE. \ref{totalresult} for a fair comparison with previous methods.

\begin{figure*}
\centering
\includegraphics[width=17cm]{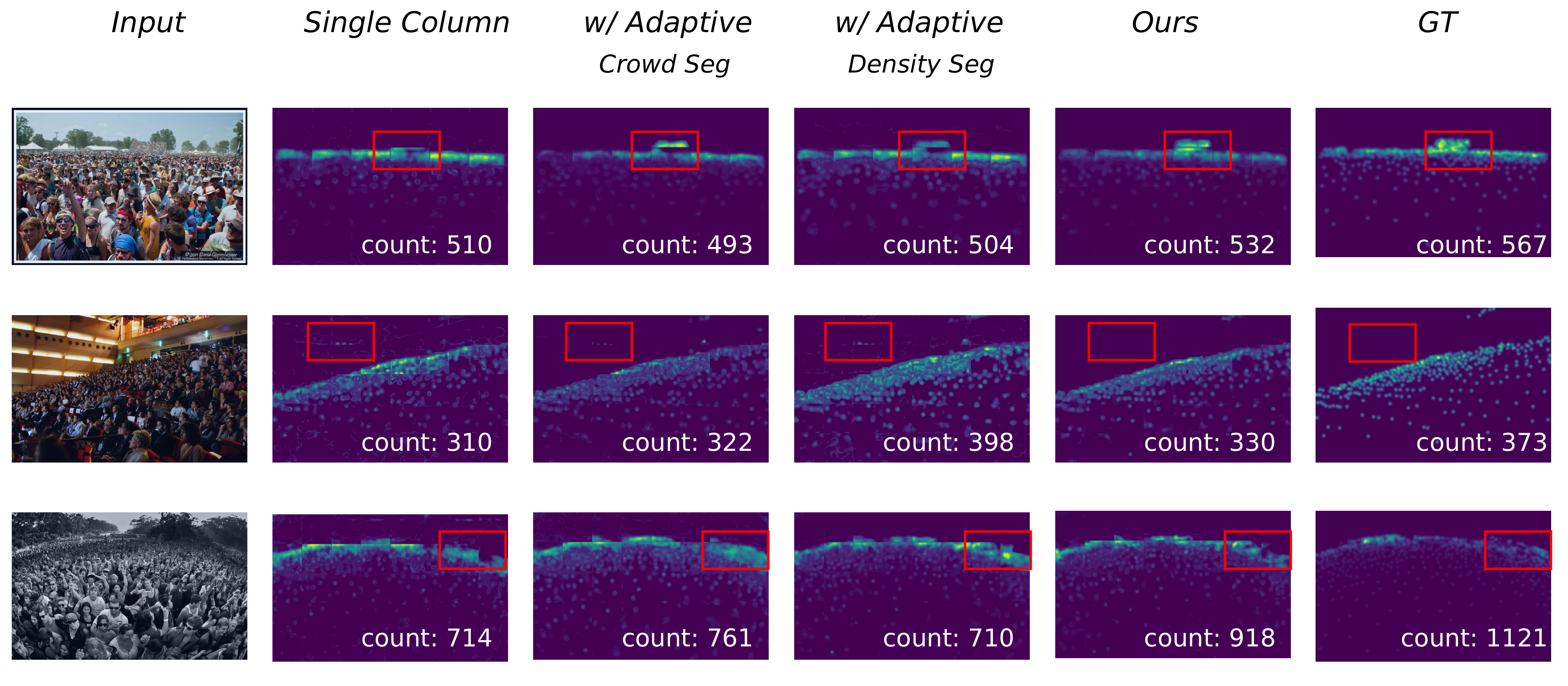}
\caption{The qualitative results of ablation studies about auxiliary tasks. The red bounding boxes are used for better visualization and comparison. \textit{Ours} and \textit{w/ Adaptive Crowd Seg} can know the crowd's spatial regions (first and third rows), and filter out the background noise (second row). On the other hand, \textit{Ours} and \textit{w/ Adaptive Density Seg} can estimate more accurate density levels across the whole density maps (second and third rows).}
\label{ablation_fig}
\end{figure*}

\noindent\textbf{Ablation on Auxiliary Tasks and Model Components.}
In this section, we evaluated the effectiveness of the auxiliary tasks, adaptively shared backbone network, and \textit{GCN}-enabled reasoning module, respectively. Please note that, in order to eliminate the performance improvement from a bigger model, we add feed-forward \textit{CNN} blocks (\(3 \times 3\) convolution with Batch Normalization) into other ablation study models in TABLE. \ref{ablationcomponent} to maintain a similar model size as ours (18.8 million parameters).
Firstly, we compared the single task density map regression network, in which we removed the GCN reasoning module, the auxiliary learning branches, and the adaptively shared backbone branches, to form a single column network structure (\textit{Single Column}).
Then we added two auxiliary branches separately and simultaneously after the single shared backbone's output to form an auxiliary learning mechanism (\textit{w/ Crowd Seg}, \textit{w/ Density Seg}, \textit{w/ Both Auxiliary}). 
To further improve the performance, we designed and added an adaptive backbone network to enable the task-shared and task-specific features being learned simultaneously (\textit{w/ Adaptive Crowd Seg}, \textit{w/ Adaptive Density Seg}, \textit{w/ Both Adaptive Auxiliary}). 
Furthermore, we evaluated the proposed \textit{GCN} reasoning module's effectiveness, which can propagate region-based density level information across the image (\textit{Ours}).
The effect of each structural component is presented in Fig. \ref{ablationcomponent}. As illustrated, the proposed auxiliary task learning mechanism (\textit{w/ Both Auxiliary}) is reduced by 14.3\% over the single-task learning method (\textit{Single Column}) via average \textit{MAE} on two datasets, the task adaptive backbone (\textit{w/ Both Adaptive Auxiliary}) reduces 6.8\% over the single shared backbone (\textit{w/ Both Auxiliary}), and the \textit{GCN} reasoning module further reduces 6.7\%. Qualitative comparison results of different modules' effectiveness in terms of predicted density maps are shown in the Fig. \ref{ablation_fig}, where the crowd segmentation auxiliary (\textit{w/ Adaptive Crowd Seg}) can help the model to focus on the features in the region of interest and filter out the background (first and second rows). On the other hand, the density level segmentation auxiliary (\textit{w/ Adaptive Density Seg}) can help to estimate more accurate density levels across the whole density map (second and third rows). We highlighted the different areas among those ablated models' density map predictions with red bounding boxes for better visualization and comparison.

Moreover, in TABLE. \ref{offloss}, we further indirectly evaluate the auxiliary tasks' effectiveness in this work. Specifically, for other ablation study models except for \textit{Ours}, we maintained the same network structure as \textit{Ours} to keep the same model size (18.8 million parameters) but switched off the two auxiliary tasks' loss functions. In TABLE. \ref{offloss}, it proves that the supervision from multi-granularity information of auxiliary tasks contributes to the final counting performance in this work. Without \(L_{CS}\) and \(L_{DS}\) losses, the counting error increases by an average of 21.75 \% on the \textit{SHA} and the \textit{JHU-Crowd} datasets via \textit{MAE}.  

\begin{table}
\centering
\caption{Ablation study results on network structure components. Each component of our network contributes to the final prediction.}
\scalebox{1.0}{
\begin{tabular}{c|cc|cc}
\hline
\multirow{2}{*}{Methds} & \multicolumn{2}{c|}{\textit{SHA}} & \multicolumn{2}{c}{\textit{JHU-Crowd}} \\ \cline{2-5} 
 & \multicolumn{1}{c}{\textit{MAE}} & \textit{RMSE} & \textit{MAE} & \textit{RMSE} \\ \hline
\textit{Single Column} & \multicolumn{1}{c}{71.3} &122.3  & 99.3 & 391.0 \\ \hline
\textit{w/ Crowd Seg} & \multicolumn{1}{c}{67.4} & 117.0 &81.6  &343.6   \\
\textit{w/ Density Seg} & \multicolumn{1}{c}{68.1} &119.9  &86.1  & 360.0  \\
\textit{w/ Both Auxiliary} & \multicolumn{1}{c}{65.2} &115.2  & 77.3 & 311.7  \\ \hline
\textit{w/ Adaptive Crowd Seg}  & \multicolumn{1}{c}{61.3} &104.6  &75.7  &300.9   \\
\textit{w/ Adaptive Density Seg} & \multicolumn{1}{c}{63.8} &108.1  &76.9  &307.8   \\
\textit{w/ Both Adaptive Auxiliary} & \multicolumn{1}{c}{60.8} &100.3  &71.9  &278.9  \\ \hline
\textit{Ours} & \multicolumn{1}{c}{\textbf{57.0}} &\textbf{98.6}  & \textbf{66.6} & \textbf{254.9}  \\ \hline
\end{tabular}
}
\label{ablationcomponent}
\end{table}

\begin{table}
\centering
\caption{Ablation study results on auxiliary tasks. Maintaining the same model structure (model size) and turning off auxiliary tasks' loss functions can implicitly prove that the auxiliary tasks contribute to the final counting.}
\scalebox{1.1}{
\begin{tabular}{c|cc|cc}
\hline
\multirow{2}{*}{Methds} & \multicolumn{2}{c|}{\textit{SHA}} & \multicolumn{2}{c}{\textit{JHU-Crowd}} \\ \cline{2-5} 
 & \multicolumn{1}{c}{\textit{MAE}} & \textit{RMSE} & \textit{MAE} & \textit{RMSE} \\ \hline
\textit{w/o \(L_{CS}\)} & \multicolumn{1}{c}{64.4} & 107.7 & 78.7 & 310.5 \\
\textit{w/o \(L_{DS}\)} & \multicolumn{1}{c}{62.0} & 104.8 & 74.9 &  302.2 \\
\textit{w/o \(L_{CS}\)} and \textit{\(L_{DS}\)} & \multicolumn{1}{c}{67.1} & 115.2 & 93.0 & 377.5 \\ \hline
Ours & \multicolumn{1}{c}{\textbf{57.0}} &\textbf{98.6}  & \textbf{66.6} & \textbf{254.9}  \\ \hline
\end{tabular}
}
\label{offloss}
\end{table}

\begin{table}
\centering
\caption{Ablation study results on the dilated rate of the proposed loss function \(L_{DCD}\). When the dilated rate is 2 and the corresponding receptive field is 5, our model can achieve the best counting performance on the \textit{SHA} and \textit{JHU-Crowd} datasets.}
\scalebox{1.1}{
\begin{tabular}{c|cc|cc}
\hline
\multirow{2}{*}{Dilated Rate} & \multicolumn{2}{c|}{\textit{SHA}} & \multicolumn{2}{c}{\textit{JHU-Crowd}} \\ \cline{2-5} 
 & \multicolumn{1}{c}{\textit{MAE}} & \textit{RMSE} & \textit{MAE} & \textit{RMSE} \\ \hline
1 & \multicolumn{1}{c}{60.1} &103.5  & 70.1 & 299.0 \\ 

3 & \multicolumn{1}{c}{58.7} &101.7  &68.7  & 288.4 \\
4 & \multicolumn{1}{c}{59.2} & 101.3 & 68.0 & 287.6 \\ \hline
2 (Ours) & \multicolumn{1}{c}{\textbf{57.0}} &\textbf{98.6}  & \textbf{66.6} & \textbf{254.9}  \\\hline

\end{tabular}
}
\label{dialtedrate}
\end{table}

\begin{table}
\centering
\caption{Ablation study results (\textit{MAE}) on our combined loss (contrastive and \textit{L2} loss), compared with single \textit{L2} loss (\textit{base}). Moreover, we applied the combined loss function to optimize previous single \textit{L2} loss based methods to demonstrate that the counting performance can be improved with the help of regional density difference based loss function \(L_{DCD}\)).}
\scalebox{1.1}
{
\begin{tabular}{c|cc|c|c}
\hline
\multirow{2}{*}{Methods} & \multicolumn{2}{c|}{\textit{SHA}} & \multicolumn{2}{c}{\textit{JHU-Crowd}} \\ \cline{2-5} 
 & \multicolumn{1}{c|}{\textit{Base}} & \textit{w/ contrastive} & \textit{Base} & \textit{w/ contrastive} \\ \hline
\textit{MCNN} \cite{zhang2016single} & \multicolumn{1}{c|}{110.2} & \textbf{108.1}  & 188.9 & \textbf{168.3} \\
\textit{CSRNet} \cite{li2018csrnet}& \multicolumn{1}{c|}{68.2} & \textbf{65.9} & 85.9 & \textbf{84.1} \\
\multicolumn{1}{c|}{\textit{CACC} \cite{liu2019context}} & \multicolumn{1}{c|}{62.3} & \multicolumn{1}{c|}{\textbf{60.8}} & \multicolumn{1}{c|}{100.1} & \multicolumn{1}{c}{\textbf{97.9}} \\ \hline 
\textit{Ours} & \multicolumn{1}{c|}{59.5} & \textbf{57.0}  & 70.8 & \textbf{66.6} \\ \hline
\end{tabular}
}
\label{ablationloss}
\end{table}

\begin{figure*}[t]
\centering
\includegraphics[width=18cm]{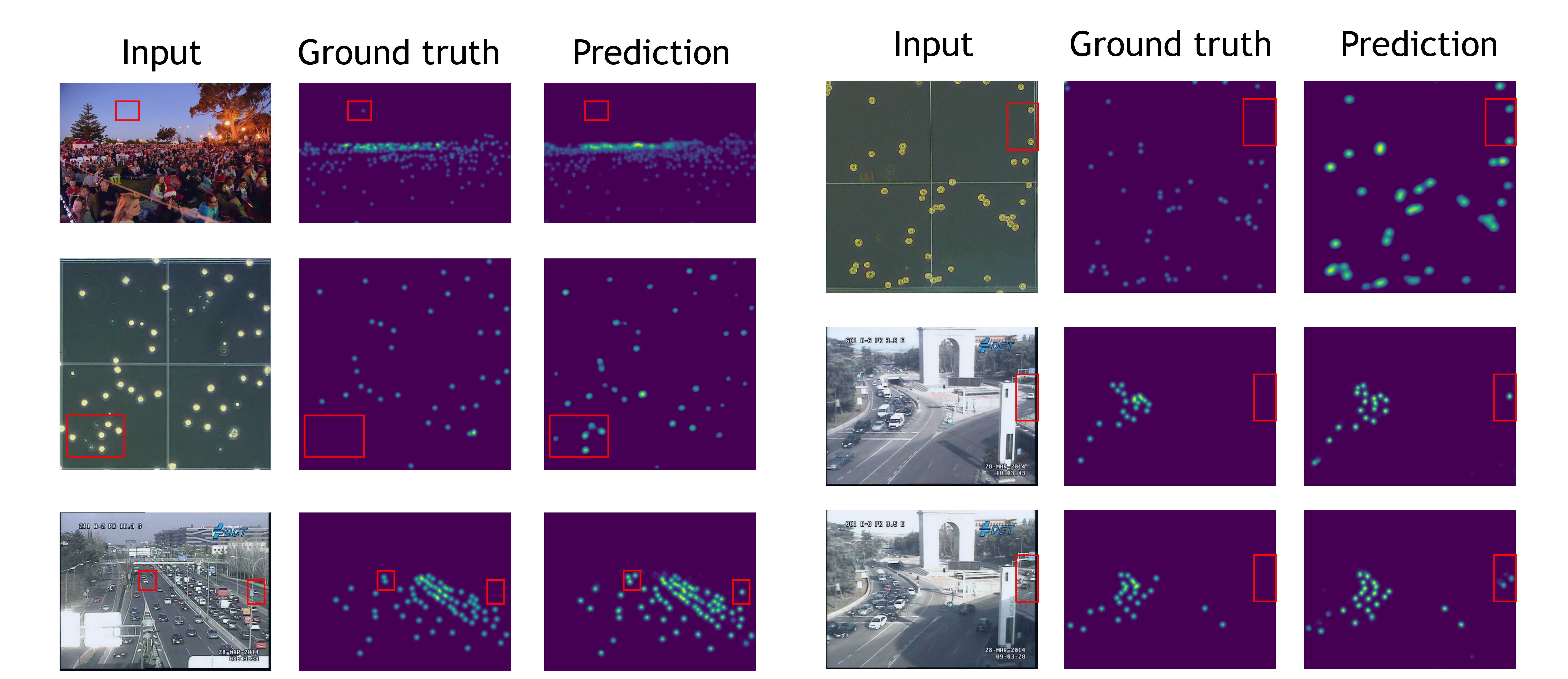} 
\caption{Comparison of our predictions and the ground truth. Our predictions are robust when there are mislabeled or wrongly labeled point annotations in the ground truth of crowd counting, cell counting, and vehicle counting datasets, respectively. The red bounding boxes are used for better visualisation and comparison.}
\label{FalseGT}
\end{figure*}

\noindent\textbf{Ablation on Loss Function.}
We performed experiments to evaluate the receptive field through different dilated rates in the proposed dilated contrastive density loss function \(L_{DCD}\). In detail, we changed the dilated rate of the \(3 \times 3\) convolution layer into 1, 2, 3, 4, which resulted in the receptive field of the \(L_{DCD}\) being like 3, 5, 7, 9. TABLE. \ref{dialtedrate} shows the comparison results; when the dilated rate is 2, our model achieves the best performance on \textit{SHA} and \textit{JHU-Crowd} datasets.

Furthermore, we conducted experiments to evaluate the effectiveness of the proposed dilated contrastive loss function, in which we removed the \(L_{DCD}\) and kept the rest of the network constant with the same trade-off hyper-parameters (\textit{Base} in TABLE. \ref{ablationloss}). Furthermore, we applied the proposed combined loss function (\textit{w/ contrastive} in TABLE. \ref{ablationloss}) into previous single \textit{L2} based methods.
We re-implemented their network with their open-source code and used the same experimental setting as our method. Fig. \ref{ablationloss} shows the comparison results of our proposed combined loss function; as illustrated, with regional density difference supervision of \(L_{DCD}\), our model attains a 3.5\% reduction compared with single \textit{L2} loss function via average \textit{MAE} on two datasets. Our proposed \(L_{DCD}\) also helps to reduce the original \textit{MCNN} \cite{zhang2016single} by 6.4\%, the \textit{CSRNet} \cite{li2018csrnet} by 2.7\%, and the \textit{CACC} \cite{liu2019context} by 2.3\% over average \textit{MAE} on two datasets. Please note that, we did not compare with other loss functions that were proposed in recent crowd counting model \cite{song2021rethinking,ma2019bayesian,wan2021generalized,wang2020distribution,wang2021uniformity,wan2020modeling}. Because those methods are not pure density map regression based methods, it is unfair to compare. 

\subsection{Discussion: Comparison with Ground Truth}\label{sec:discussion}
The underlying labeling errors (noisy ground truth) exist in most datasets due to the human annotators' errors. However, a robust model can omit the noise ground truth during training and produce a more accurate prediction.
This section showed that our model could indicate some mislabeled or wrongly labelled point annotations of the ground truth in the test dataset. This highlights the generalizability of our approach and the potential issue of the imperfect ground truth in object counting applications. Fig. \ref{FalseGT} shows a wrongly labelled point annotation (top left) case of the crowd counting test dataset, and the other cases are mislabeled point annotation of vehicle and cell counting test dataset. We highlighted the wrongly labelled or mislabeled area with red bounding boxes for better visualization and comparison.

\section{Conclusion}
We proposed a novel framework for auxiliary task learning based counting by employing an adaptively shared backbone, a \textit{GCN} reasoning module and a novel dilated contrastive density loss function. 
Our model advocates task-shared and task-specified features to be learned simultaneously. The proposed method highlights that cross-domain reasoning in graph through \textit{GCNs} using crowd segmentation and density level segmentation can significantly improve feature learning in density map regression tasks. With our proposed loss function's regional density difference supervision, our model set a new state-of-the-art among auxiliary task learning based counting methods on seven challenging benchmarks.


%




\ifCLASSOPTIONcaptionsoff
  \newpage
\fi



%

\bibliographystyle{IEEEtran}
\bibliography{IEEEfull}

\end{document}